# Automated Detection of Pre-Disaster Building Images from Google Street View

Chul Min Yeum[a], Ali Lenjani[b], Shirley J. Dyke[a,b], Ilias Bilionis[b]

[a] Lyles School of Civil Engineering, Purdue University, IN, 47906, United States.
[b] School of Mechanical Engineering, Purdue University, IN, 47906, United States.

## Extended Abstract

After a disaster, teams of structural engineers collect vast amounts of images from damaged buildings to obtain lessons and gain knowledge from the event. Images of damaged buildings and components provide valuable evidence to understand the consequences on our structures. However, in many cases, images of damaged buildings are often captured without sufficient spatial context. Also, they may be hard to recognize in cases with severe damage. Incorporating past images showing a pre-disaster condition of such buildings is helpful to accurately evaluate possible circumstances related to a building's failure.

One of the best resources to observe the pre-disaster condition of the buildings is Google Street View [1]. A sequence of 360° panorama images which are captured along streets enables all-around views at each location on the street. Once a user knows the GPS information near the building, all external views of the building can be made available.

In this study, we develop an automated technique to extract past building images from 360° panorama images (panoramas) serviced by Google Street View. Users only need provide a geo-tagged image, collected near the target building, and the rest of the process is fully automated. High-quality and undistorted building images are extracted from past panoramas. Since the panoramas are collected from various locations near the building along the street, the user can identify its pre-disaster conditions from the full set of external views.

Overall, the developed technique is divided into four main steps: In Step 1, as a preliminary step, we build an object detector using a large volume of ground-truth building images to extract the buildings on images. We use a region-based convolutional neural network algorithm to train a robust building detector [4]. This step is conducted only one time for extracting several building images from the panoramas. Step 2 is to download the panoramas near the target building from Google Street View [1]. The approximate GPS location of the target building is obtained from the geo-tagged image that a user provides. In Step 3, the true building location in the GPS coordinate is computed using the geometric relationship between the building and the panoramas. The location of the building facilitates estimating the optimal projection direction of each panorama. By projecting each panorama with the corresponding direction, 2D images are generated, which contain a high-quality and undistorted building view. Finally, in Step 4, the trained object detector is applied to detect and localize the building in each of the 2D images.

To demonstrate the technique developed, the house in Figure 1(a) is selected as our target building, which is located in Rockport, Texas, United States. This house has significant damage during Hurricane Harvey in 2017 [2]. The geo-tagged image in Figure 1(a) was collected in an actual post-disaster reconnaissance mission [3] and is used as the input for the technique. First, a classifier for detecting houses is trained using a large volume of house images collected from the same geographical region. Faster region-based convolutional neural network, implemented as Python, is used for training this classifier [4]. Next, the three nearest high-resolution panoramas to the location of the input image are downloaded from Google Street View in Figure 1(b). Since these panoramas contain the 360° spherical region in the 2D image, the house in each panorama is highly distorted, as shown in Figure 1 (b). To remove distortion of the house view in the panorama, the 2D image is projected from each of the panorama using the optimal projection direction. This makes the house appear in the center of each image with good quality in Figure 1(c). Finally, the trained house classifier is applied to the 2D images in Figure 1(c) to extract the only house region. In Figure 1(d), the house images are successfully extracted and contain the primary face of the house.

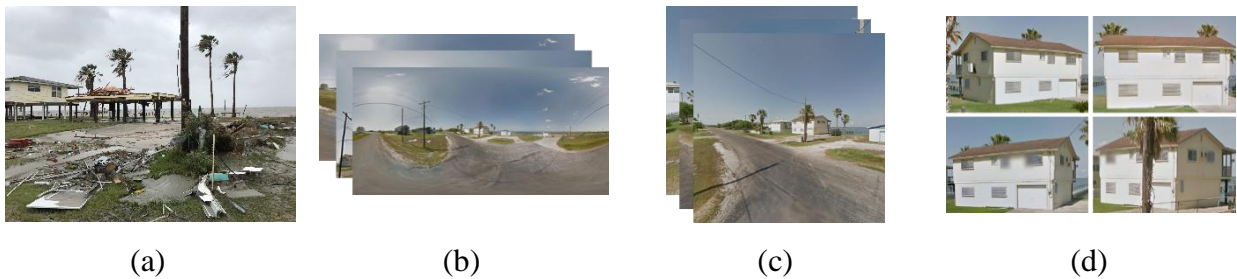

(a)            (b)            (c)            (d)

Figure 1. Detection of a house from Google Street View: (a) post-disaster building images collected after Hurricane Harvey in 2017, (b) 360° panorama images downloaded from Google Street View, (c) 2D images projected from the panorama images using the optimal project directions, and (d) house images detected from the images in (c).